\documentclass{article}

\usepackage[preprint]{corl_2023} % Uncomment for pre-prints (e.g., arxiv); This is like ``final'', but will remove the CORL footnote.

\usepackage{graphics}
\usepackage[utf8]{inputenc}
\usepackage{graphicx}
\usepackage{booktabs}
\usepackage{amsmath} % assumes amsmath package installed
\usepackage{amssymb}  % assumes amsmath package installed
\usepackage{amsfonts}
\usepackage{mathtools}
\usepackage[font=small]{caption}
\usepackage[font=small]{subcaption}
\usepackage{booktabs}
\usepackage{wrapfig}
\usepackage{flushend}
\usepackage{pifont}% http://ctan.org/pkg/pifont
\usepackage{arrayjob}

\usepackage{etoolbox}
\usepackage[capitalise]{cleveref}
\usepackage{pgfplots}
\usepgfplotslibrary{groupplots,dateplot}
\usetikzlibrary{patterns,shapes.arrows, calc}
\pgfplotsset{compat=newest}
\usepackage{hyperref}
\usepackage{enumitem}

\pgfplotsset{
    every axis/.append style={
        line width=1pt,
        tick style={line width=0.8pt}},
    label style={font=\small},
    tick label style={font=\footnotesize}
}

\renewrobustcmd{\bfseries}{\fontseries{b}\selectfont}
\renewrobustcmd{\boldmath}{}
\newrobustcmd{\B}{\bfseries}
\usepackage{changes}

\usepackage{siunitx}
\usepackage{bm}

\newcommand{\R}{\mathbb{R}}

\DeclareMathAlphabet\mathbfcal{OMS}{cmsy}{b}{n}

\DeclarePairedDelimiterX{\infdivx}[2]{(}{)}{%
  #1\;\delimsize|\delimsize|\;#2%
}

\def\BibTeX{{\rm B\kern-.05em{\sc i\kern-.025em b}\kern-.08em
    T\kern-.1667em\lower.7ex\hbox{E}\kern-.125emX}}

\usepackage{caption}
\usepackage{subcaption}
\usepackage{booktabs}
\usepackage{xcolor}
\usepackage{colortbl}
\usepackage{pgfplots}
\pgfplotsset{compat=1.16}
\usetikzlibrary{intersections}
\usepgfplotslibrary{fillbetween}
\definecolor{lightblue}{RGB}{173,216,230} % Define light blue color
\definecolor{lightred}{RGB}{255,182,193}
\usepackage{grffile}
 
\title{LOPR: Latent Occupancy PRediction using Generative Models\\[0.5em]
\large\textcolor{red}{\textit{*We recommend referring to the peer-reviewed and updated version of this approach, available at \href{https://arxiv.org/abs/2407.21126}{https://arxiv.org/abs/2407.21126}.}}}

\author{
  Bernard Lange, Masha Itkina, and Mykel J.~Kochenderfer\\
  Department of Aeronautics and Astronautics\\
  Stanford University \\
  United States\\
  \texttt{\{blange, mitkina, mykel\}@stanford.edu} \\
}

\begin{document}
\maketitle

%===============================================================================
\vspace{-0.4cm}
\begin{abstract}
Environment prediction frameworks are integral for autonomous vehicles, enabling safe navigation in dynamic environments. LiDAR generated occupancy grid maps (L-OGMs) offer a robust bird's eye-view scene representation that facilitates joint scene predictions without relying on manual labeling unlike commonly used trajectory prediction frameworks. Prior approaches have optimized deterministic L-OGM prediction architectures directly in grid cell space. While these methods have achieved some degree of success in prediction, they occasionally grapple with unrealistic and incorrect predictions.  %addressing significant challenges in current trajectory prediction models.
We claim that the quality and realism of the forecasted occupancy grids can be enhanced with the use of generative models. We propose a framework that decouples occupancy prediction into: representation learning and stochastic prediction within the learned latent space. 
Our approach allows for conditioning the model on other available sensor modalities such as RGB-cameras and high definition maps. We demonstrate that our approach achieves state-of-the-art performance and is readily transferable between different robotic platforms on the real-world NuScenes and Waymo Open Perception datasets.
%This unique capability also positions L-OGM prediction as a promising unsupervised pre-training task for autonomous driving. 

%Environment prediction frameworks are integral for autonomous vehicles, enabling safe navigation in dynamic environments. Prior approaches have used occupancy grid maps as bird's eye-view representations of the scene, optimizing prediction architectures directly in grid cell space. While these methods have achieved some degree of success in spatiotemporal prediction, they occasionally grapple with unrealistic and incorrect predictions. 
%We claim that the quality and realism of the forecasted occupancy grids can be enhanced with the use of generative models. We propose a framework that decouples occupancy prediction into two parts: representation learning and stochastic prediction within the learned latent space. 
%Our approach allows for conditioning the model on other commonly available sensor modalities such as RGB-cameras and high definition~(HD) maps. We demonstrate that our approach achieves state-of-the-art performance and is readily transferable between different robotic platforms on the real-world NuScenes, Waymo Open, and our custom robotic datasets.
\end{abstract}


%===============================================================================

\section{Introduction}
\label{sec:intro}
% Why environment prediction? 
Accurate environment prediction algorithms are essential for autonomous vehicle (AV) navigation in urban settings.
Experienced drivers understand scene semantics and recognize the intents of other agents to anticipate their trajectories and safely navigate to their destination. To replicate this process in AVs and other robotic platforms, many environment prediction approaches have been proposed, employing different environment representations and modeling assumptions~\citep{itkina2019dynamic, toyungyernsub2021double, lange2020attention, salzmann2020trajectron++, ivanovic2020mats, chai2019multipath, tang2019multiple, mahjourian2022occupancy, ngiam2021scene}. 

%%%%%%
% Object-based representation

The modern AV stack comprises several sequential modules trained independently on labeled data. For environment reasoning, object-based prediction algorithms are commonly used ~\cite{chai2019multipath, tang2019multiple, salzmann2020trajectron++, zhao2020tnt}, which rely on perception systems to create a vectorized representation of the scene with pre-defined agents and environmental features. However, this approach has multiple limitations. 1) It generates marginalized future trajectories for individual agents, rather than a holistic scene prediction which complicates the integration with planning modules~\cite{chen2022scept}. % with interactions between predicted agent. 
%It complicates the integration with planning and control modules, often necessitating additional rule-based methods to produce the complete scene prediction~\cite{chen2022scept}. 
2) Their reliance on labeled data, sourced either manually or from off-board perception systems~\cite{ettinger2021large, qi2021offboard}, diverges from on-board noisy detections. 3) They do not take any sensor measurements into account and depend solely on object detection algorithms which can fail in suboptimal conditions~\citep{delecki2022we, dreissig2023survey}. These approaches also exclude social and topological cues that humans naturally perceive, emphasizing the importance of end-to-end perception-prediction learning~\cite{nayakanti2022wayformer}. The current drawbacks render the AV stack susceptible to cascading failures, and can lead to poor generalization to unforeseen scenarios. These limitations underscore the need for alternative end-to-end, self-supervised environment modeling approaches.
%Compared to object-based prediction,

Given these challenges, occupancy grid maps generated from LiDAR measurements (L-OGMs) have gained popularity as a form of scene representation for prediction. This popularity is due to their minimal data preprocessing requirements, eliminating the need for manual labeling, ability to model the joint prediction of the scene with an arbitrary number of agents (including interactions between agents), and robustness to partial observability and detection failures~\cite{itkina2019dynamic, mohajerin2019multi, lange2020attention, toyungyernsub2021double}.
In addition, the sole requirement for their deployment is a LiDAR sensor, simplifying transfer between different platforms. Our focus is on end-to-end ego-centric L-OGM prediction generated using uncertainty-aware occupancy state estimation approaches~\citep{elfes1989using}. Due to its generality and ability to scale with unlabeled data, we hypothesize that such an L-OGM prediction framework could also serve as a pre-training objective, i.e. a foundational mode, for supervised tasks such as trajectory prediction.

\begin{figure}[t]
  \begin{minipage}[b]{0.4\textwidth}
    \centering
    \includegraphics[width=4.0cm]{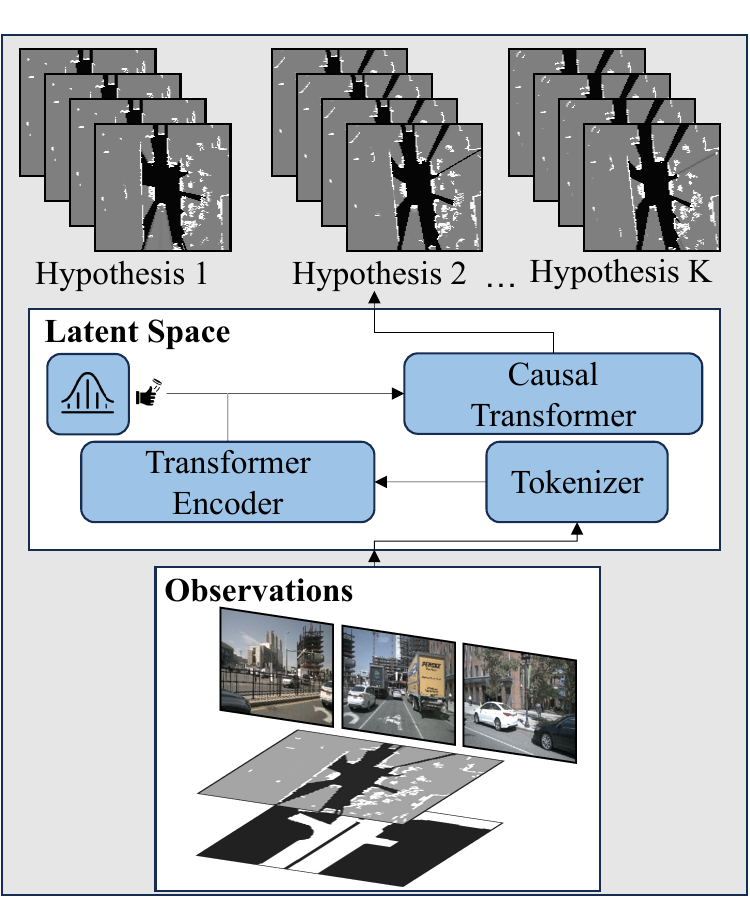}
    %\caption*{LOPR Visualization}
    \label{fig:figure-label}
  \end{minipage}%
  % \hfill
  \begin{minipage}[b]{0.60\textwidth} % Adjust the width as desired
    \centering
    \tiny
    \begin{tabular}{@{}p{2.15cm}p{0.7cm}p{0.2cm}p{0.2cm}p{0.4cm}p{0.5cm}p{1.0cm}c@{}} % Remove the padding around the table
      \toprule
      Method & Rep. & Maps & Cam. & Partial obs. & Stochast. & Prediction Type \\
      \midrule
      \citet{chai2019multipath} & \textcolor{red}{Vector} & \textcolor{green}{\checkmark} & \textcolor{red}{\ding{55}} & \textcolor{red}{\ding{55}} & \textcolor{green}{GMM} & \textcolor{red}{Per-agent} \\
      \citet{ivanovic2020mats} & \textcolor{red}{Vector} & \textcolor{green}{\checkmark} & \textcolor{red}{\ding{55}} & \textcolor{red}{\ding{55}} & \textcolor{green}{GMM} & \textcolor{red}{Per-agent} \\
      \citet{gu2021densetnt} & \textcolor{red}{Vector} & \textcolor{green}{\checkmark} & \textcolor{red}{\ding{55}} & \textcolor{red}{\ding{55}} & \textcolor{green}{Goal} & \textcolor{red}{Per-agent} \\
      \citet{nayakanti2022wayformer} & \textcolor{red}{Vector} & \textcolor{green}{\checkmark} & \textcolor{red}{\ding{55}} & \textcolor{red}{\ding{55}} & \textcolor{green}{GMM} & \textcolor{red}{Per-agent} \\
      \citet{shi2022motion} & \textcolor{red}{Vector} & \textcolor{green}{\checkmark} & \textcolor{red}{\ding{55}} & \textcolor{red}{\ding{55}} & \textcolor{green}{GMM} & \textcolor{red}{Per-agent} \\
      \citet{itkina2019dynamic} & \textcolor{green}{L-OGM} & \textcolor{red}{\ding{55}} & \textcolor{red}{\ding{55}} & \textcolor{green}{\checkmark} & \textcolor{red}{\ding{55}} & \textcolor{green}{Scene} \\
      \citet{lange2020attention} & \textcolor{green}{L-OGM} & \textcolor{red}{\ding{55}} & \textcolor{red}{\ding{55}} & \textcolor{green}{\checkmark} & \textcolor{red}{\ding{55}} & \textcolor{green}{Scene} \\
      % \rowcolor{gray!15}
      \citet{toyungyernsub2022dynamics} & \textcolor{green}{L-OGM} & \textcolor{red}{\ding{55}} & \textcolor{red}{\ding{55}} & \textcolor{green}{\checkmark} & \textcolor{red}{\ding{55}} & \textcolor{green}{Scene} \\
      \citet{mahjourian2022occupancy} & \textcolor{red}{V-OGM} & \textcolor{green}{\checkmark} & \textcolor{red}{\ding{55}} & \textcolor{red}{\ding{55}} & \textcolor{red}{\ding{55}} & \textcolor{green}{Scene} \\
      \citet{mersch2022self} & \textcolor{green}{PCL} & \textcolor{red}{\ding{55}} & \textcolor{red}{\ding{55}} & \textcolor{green}{\checkmark} & \textcolor{red}{\ding{55}} & \textcolor{green}{Scene} \\
      \citet{wu2020motionnet} & \textcolor{green}{PCL} & \textcolor{green}{\checkmark} & \textcolor{green}{\checkmark} & \textcolor{green}{\checkmark} & \textcolor{red}{\ding{55}} & \textcolor{green}{Scene} \\
      \midrule
      LOPR (ours) & \textcolor{green}{L-OGM} & \textcolor{green}{\checkmark} & \textcolor{green}{\checkmark} & \textcolor{green}{\checkmark} & \textcolor{green}{Variat.} & \textcolor{green}{Scene} \\
      \bottomrule
    \end{tabular}
    \caption*{Representations and prediction types in common approaches}
    \label{tab:table-label}
  \end{minipage}
\caption{(Left) Latent Occupancy PRediction (LOPR). We decouple the prediction task into task-independent representation learning, and task-dependent prediction in the latent space. (Right) Comparison with other approaches in terms of representation type, sensors, stochasticity assumptions, and prediction type. Only LOPR makes stochastic predictions of the scene conditioned on all sensors without the need for manually labelled data.}
\label{fig:overview}
\vspace{-1.0em} % Add some vertical spacing between the rows
\end{figure}

The task of OGM prediction is typically approached similarly to video prediction, by framing the problem as self-supervised sequence-to-sequence learning. In this approach, a scenario is dissected into a history sequence and a target prediction sequence. ConvLSTM-based architectures~\citep{xingjian2015convolutional} have been used in previous work for this task due to their ability to handle the spatiotemporal representation of inputs and outputs~\citep{itkina2019dynamic, schreiber2019long, lange2020attention, toyungyernsub2021double, schreiber2021dynamic}. These approaches are optimized end-to-end in grid cell space, do not account for the stochasticity present in the scene, and neglect other available sensor modalities, e.g. RGB cameras and high definition (HD) maps. As a result, they suffer from blurry predictions, especially at longer time horizons. We propose a prediction framework that reasons over potential futures in the latent space of generative models. It is trained on sensor modalities such as L-OGMs, 2D RGB cameras, and maps without the need for manual labeling. We illustrate our framework in \cref{fig:overview} and compare it with other methods.

% \begin{figure}[t]
%     \centering
%     \includegraphics[height=2cm, width=13cm]{example-image-a}
%     \caption{Latent Occupancy PRediction (LOPR). We decouple the OGM prediction task into two stages: task-independent representation learning, and task-dependent prediction.}
%     \label{fig:overview}
% \end{figure}

Recent work has shown generative models can produce high-quality~\citep{goodfellow2014generative, karras2020analyzing} and controllable~\citep{kingma2013auto, larsen2016autoencoding, higgins2016beta} samples. In robotics, generative models have been used to find compact representations of images in planning~\citep{ha2018recurrent,xie2020learning, schwarting2021deep}, control~\citep{lenz2015deepmpc, wahlstrom2015pixels, xu2020learning}, and simulation~\citep{kim2021drivegan}. We claim that generative models are similarly capable of accurately encoding and decoding L-OGMs, alongside providing a controllable latent space for high-quality predictions. We employ a generative model to learn a low-dimensional latent space, which encodes the features needed to generate realistic predictions and makes use of available input modalities, such as L-OGM, RGB camera, and map-based observations. We then train a stochastic prediction network in this latent space to capture the dynamics of the scene.

Existing object-based methods use a vectorized representation to predict trajectories~\citep{chai2019multipath, ivanovic2020mats, gu2021densetnt} or vectorized OGMs (V-OGMs)~\cite{mahjourian2022occupancy}, overlooking important perceptual cues in their predictions. Prior L-OGM-based works~\cite{itkina2019dynamic, lange2020attention, mersch2022self} do not use available sensor modalities, and consider only deterministic predictions. Our framework addresses these weaknesses in the following contributions:

%due to the difficulty of modeling stochasticity for high-dimensional spatiotemporal inputs.
% \setlist{nosep}
\begin{itemize}[itemsep=0pt,parsep=0pt] %[noitemsep]
\item We introduce a framework named \textbf{L}atent \textbf{O}ccupancy \textbf{PR}ediction (LOPR), which performs stochastic L-OGM prediction in the latent space of a generative architecture conditioned on other available sensor modalities, like RGB cameras and maps.
\item Through experiments on NuScenes~\citep{caesar2020nuscenes} and the Waymo Open Dataset~\citep{sun2020scalability}, we show that LOPR outperforms SOTA OGM prediction methods qualitatively and quantitatively. 
\item We demonstrate that LOPR can be conveniently transferred between different robotic platforms.
\end{itemize}

\section{Related Work}
\label{sec:related}
\textbf{OGM Prediction}:
The majority of prior work in OGM prediction generates OGMs with LiDAR measurements (L-OGMs) and uses an adaptation on the recurrent neural network (RNN) with convolutions. \citet{dequaire2018deep} applied a Deep Tracking approach~\citep{ondruska2016deep} to track objects through occlusions and predict future binary OGMs with an RNN and a spatial transformer~\cite{jaderberg2015spatial}. \citet{schreiber2019long} provided dynamic occupancy grid maps (DOGMas) with cell-wise velocity estimates as input to a ConvLSTM~\cite{xingjian2015convolutional} for environment prediction from a stationary platform. \citet{schreiber2021dynamic} extended this work to forecast DOGMas in a moving ego-vehicle setting. \citet{mohajerin2019multi} applied a difference learning approach to predict OGMs as seen from the coordinate frame of the first observed time step. \citet{itkina2019dynamic} used the PredNet ConvLSTM architecture~\cite{lotter2016deep} to achieve ego-centric OGM prediction. \citet{lange2020attention} reduced the blurring and the gradual disappearance of dynamic obstacles in the predicted grids by developing an attention augmented ConvLSTM mechanism. Concurrently, \citet{toyungyernsub2021double} addressed obstacle disappearance with a double-prong framework assuming knowledge of the static and dynamic obstacles. Predicted OGMs often lack agent identity information. \citet{mahjourian2022occupancy} addressed this by exploiting occupancy flow estimates to trace back the agent identities from the observed OGM frames. They used upstream perception detections to generate OGMs with vectorized represenations (V-OGMs) and hence require manual labeling. Unlike prior work, we perform stochastic OGM predictions in the latent space of generative models of all available sensors without any manual labeling.

\textbf{Representation Learning in Robotics}: The objective of representation learning is to identify a low-dimensional and disentangled representation that makes it easier to achieve the desired performance on a task. Many robotics applications are inspired by the seminal papers on the autoencoder~(AE)~\cite{baldi1989neural, hinton1990connectionist, hinton1993autoencoders}, the variational autoencoder~(VAE)~\cite{kingma2013auto}, and the generative adversarial network~(GAN)~\cite{goodfellow2014generative}. \citet{ha2018recurrent} proposed a World Model, where they used a VAE to compress observations and maximize the expected cumulative reward. \citet{kim2021drivegan} applied the World Model to neural network simulation for autonomous driving, where they merged the VAE~\citep{kingma2013auto} and StyleGAN~\citep{karras2020analyzing} to increase fidelity of the generated scenes. Similarly, latent spaces have been used in a plethora of other planning and control approaches to learn latent dynamics from pixels~\citep{wahlstrom2015pixels, buesing2018learning, pmlr-v97-hafner19a, gelada2019deepmdp, xu2020learning}, generate fully imagined trajectories~\citep{hafner2019dream}, model multi-agent interactions~\citep{xie2020learning}, learn competitive policies through self-play~\citep{schwarting2021deep}, imagine goals in goal-conditioned policies~\citep{kurutach2018learning, anand2019unsupervised}, and perform meta- and offline reinforcement learning~\citep{zhao2020meld,zhou2020plas}. Diffusion-based generative models~\citep{sohl2015deep, ho2020denoising, song2020denoising, ho2022video} are increasingly gaining traction due to their high sampling quality. However, their computationally intensive sampling process poses a significant obstacle for real-time robotic applications in the context of perceptual generation. In video prediction tasks, variational and adversarial components have been incorporated into the architecture to capture data stochasticity~\citep{babaeizadeh2017stochastic} and improve the realism of the forecasted frames~\citep{lee2018stochastic}, respectively. Since then, large-scale architectures (over 300 million parameters)~\cite{weissenborn2019scaling, yu2023magvit} have been developed for general video prediction, which are not real-time capable on robot hardware.
We present the first method, to our knowledge, that performs stochastic L-OGM prediction with transformers entirely in the latent space of a generative model while remaining real-time feasible and parameter efficient (less than 4 million parameters). 
% Contrary to our approach, none of these models perform OGM prediction entirely in the latent space of generative models.
\section{LOPR: Latent Occupancy PRediction}

\begin{figure}[t]
    \centering
    \includegraphics[width=13.5cm]{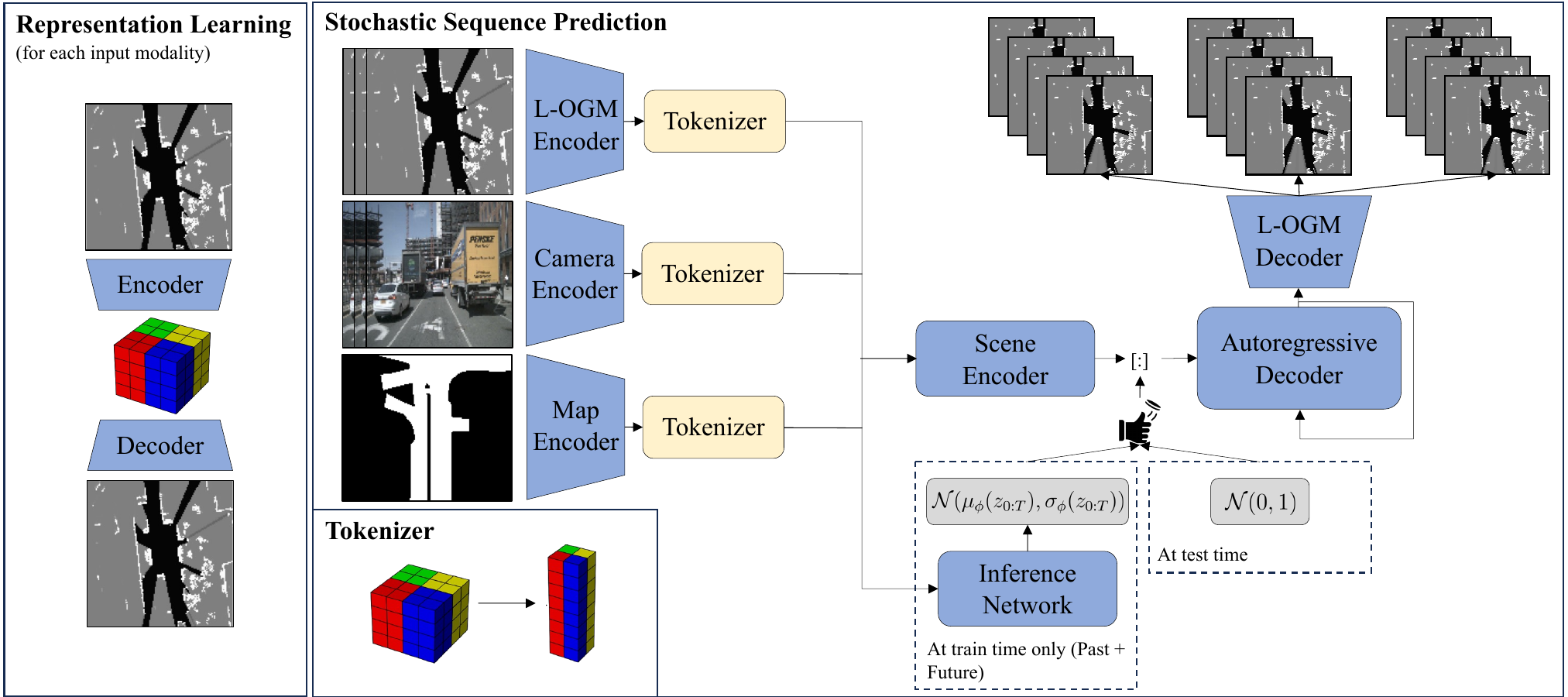}
  \caption{The illustration shows the LOPR framework which consists of (1) representation learning and (2)~stochastic sequence prediction. In the representation learning stage, we train an encoder and decoder in an unsupervised manner. In the sequence prediction stage, we convert our OGM dataset to the low-dimensional representation, and perform training entirely in the latent space of our pre-trained generative model.}   
  \label{fig:LOPR_details}
  \vspace{-1.2em}
\end{figure}

We propose the Latent Occupancy PRediction (LOPR) model, a framework designed to generate stochastic scene predictions in the form of OGMs. The model uses representations provided by sensor modalities, such as LiDAR-generated OGMs, RGB cameras, and maps. It does not require any manually labeled data and can be deployed on any robot equipped with at least a LiDAR sensor. A visualization of the framework is provided in \cref{fig:LOPR_details}. The model separates the prediction task into (1) learning the environment representation and (2) making predictions in the latent space of a generative model. In the representation learning phase, a VAE-GAN is trained to acquire a pre-trained latent space of rasterized sensor measurements. During the prediction stage, an auto-regressive transformer~\citep{vaswani2017attention} network is trained within the pre-trained latent space to predict future OGMs. It operates over patches of each latent vector to reduce the dimensionality of the prediction network and employs a series of auxiliary tasks incorporating various sequence masking strategies during training to further improve performance. 

% Its task is to ensure that the generated OGMs are temporally and semantically consistent. 
%The encoder learns a low-dimensional representation for OGMs which can later be used for prediction, and the generator decodes the latent space to OGM space.
% For the environment representation, we use OGMs generated from LiDAR measurements following the procedure described by~\citet{itkina2019dynamic}. 
%In the following subsections, we detail the unsupervised representation learning and supervised learning stages, respectively.

\subsection{Representation Learning}
\label{sec:unsupervied}

In the first stage of training, we acquire a pre-trained latent space of all input modalities by training an encoder $\mathcal{E}$ and a decoder $\mathcal{D}$. Given the input modality $x\in\R ^{C \times W \times H}$, the encoder outputs a low-dimensional latent vector $z \in\R ^{c \times h \times w}$, and the decoder maps the latent vector to a reconstruction $\hat{x}\in\R ^{C \times W \times H}$. The framework is trained using a combination of perceptual loss~\citep{zhang2018unreasonable}, Kullback-Leibler (KL) regularization~\citep{kingma2013auto}, patch-based adversarial losses~\citep{isola2017image} and path regularization~\cite{karras2020analyzing}:
\begin{equation}
\label{eqn:autoenc}
    L_{\text {VAEGAN}}=\min _{\mathcal{E}, \mathcal{D}} \max _\psi\left(L_\text{LPIPS}(x, \mathcal{D}(\mathcal{E}(x)))-\gamma L_\text{adv}(\mathcal{D}(\mathcal{E}(x)))+\beta L_\text{KL}(x ; \mathcal{E}, \mathcal{D}) + L_\text{reg} \right).
\end{equation}
We employ the adversarial loss to increase the visual fidelity of of the generated samples, KL regularization to encourage the posterior $q(z \mid x)$ to be clustered close the prior $p(z) = \mathcal{N}(0,I)$. %We find the KL regularization to be critical for good performance in the downstream prediction task. 
%We analyze the performance of the pre-trained latent space in \cref{sec:latent_analysis}. %and provide implementation details in \cref{sec:representation_learning_details}.

\subsection{Stochastic Sequence Prediction}
Given the pre-trained latent space of our sensor data, we train a stochastic sequence prediction network that receives a history of observations and outputs a distribution of potential future scenarios $p_{\theta}(z_{P:T} \mid z_{0:P})$, where $z_{0:P}$ represents compressed observations over $P$ timesteps, $T$ signifies the total sequence length, and $\theta$ are the network weights. To simplify notation, we assume an abuse of notation in that when $z_t$ corresponds to an observation, it includes all sensor modalities; conversely, when referring to the future, it includes only the OGM representation. The environment prediction task is inherently multimodal, and the latent vectors contributing to this stochasticity are unobservable. Drawing from prior work on video prediction~\citep{babaeizadeh2017stochastic}, we introduce a latent vector $z_{stoch} \sim p(z_{stoch})$ to encapsulate this stochasticity and extend our model to $p_{\theta}(z_{P:T} \mid z_{0:P}, z_{stoch})$. During training, we extract the true posterior $p(z_{stoch} \mid z_{0:T})$ using an inference network $z_{stoch} \sim q_{\phi}(z_{stoch} \mid z_{0:T})$, while at test time, we sample from a pre-defined prior $p(z_{stoch})$. The framework is optimized using the variational lower bound~\cite{kingma2013auto}:
\begin{equation}
\label{eqn:elbo}
\mathcal{L}(\mathbf{z})=-\mathbb{E}{q_\phi\left(z_{stoch} \mid z_{0: T}\right)}\left[\log p_\theta\left(z_{P: T} \mid z_{0: P}, z_{stoch}\right)\right]+D_{K L}\left(q_\phi\left(z_{stoch} \mid z_{0: T}\right) | p(z_{stoch})\right),
\end{equation}

where $D_{KL}$ is the Kullback-Leibler divergence, and $p(z_{stoch})$ is standard Gaussian~\cite{babaeizadeh2017stochastic, denton2018stochastic, lee2018stochastic, franceschi2020stochastic}.

We implement our model using a transformer-based architecture, comprised of a scene encoder $\mathcal{P}_{encoder,\theta}$, an inference network $\mathcal{Q}_{\phi}$, and an autoregressive decoder $\mathcal{P}_{decoder,\theta}$. In all transformation-based operations, positional information is provided through a sinusoidal positional encoding~\citep{vaswani2017attention}. To reduce the number of parameters required in the attention layer for processing each token and to facilitate spatial attention, each latent vector $z \in \R^{c \times h \times w}$ is reshaped along the spatial dimensions to a sequence of smaller cuboids $z \in \R^{k^2c \times \frac{h}{k} \times \frac{w}{k}}$. This is followed by spatial partitioning. A single latent vector $z$ thus generates a sequence of $\frac{hw}{k^2}$ tokens, each of length $k^2c$. 

% It allows to Transformer to attend spatially within the same timestep and reduces the overall number of parameters required to process the token in the attention layer. 

The scene encoder and inference network are implemented using a transformer architecture, wherein each token can attend to every other token. The scene encoder takes in tokens generated with latent vectors for all sensor modalities and outputs scene contexts $scene_{emb}\in\R^{kc^2}$. The posterior network takes in the all OGM latent vectors, including the future ones, and outputs parameters for the Gaussian distribution $\mu$ and $\sigma$, used for $z_{stoch}$:
\begin{align}
scene_{emb} &= \mathcal{P}_{encoder,\theta}(z_{0:P}) \\
z_{stoch} &\sim \mathcal{N}\left(\mu_\phi\left(z_{0: T}\right), \sigma_\phi\left(z_{0: T}\right)\right).
\label{eqn:prediction_encoder}
\end{align}

Both $scene_{emb}$ and $z_{stoch}$ are concatenated along the temporal dimension and serve as memory tokens to $\mathcal{P}_{decoder,\theta}$, which is implemented using a causal transformer to generate predictions:
\begin{align}
z_{t} = \mathcal{P}_{decoder,\theta}(z_{0:t-1}, mem= \left[scene_{emb}, z_{stoch}\right])
\label{eqn:prediction_decoder}
\end{align}

In the final step, the predicted compressed representations are concatenated, reshaped back to their original dimensions, and then fed into the pretrained decoder $\mathcal{D}$ to produce the OGM predictions. %\textcolor{blue}{We provide more architectural details in the appendix.} 
\section{Experiments}
\label{sec:experiments}
We analyze our proposed framework based on its pre-trained latent space and how well it performs on environment prediction tasks. We also consider how additional sensor modalities affect the prediction quality. Our framework is tested against prior methods in L-OGM forecasting and top-performing sequence-to-sequence video prediction models of similar scale. We use the open-source datasets, NuScenes \cite{caesar2020nuscenes} and the Waymo Open Dataset~\citep{sun2020scalability}, for training and evaluation. Our results show that LOPR reaches SOTA performance in the prediction task. Our code with additional visualizations is available \href{https://github.com/sisl/LOPR}{here}.

\subsection{Datasets}
\label{sec:datasets}

\textbf{NuScenes Dataset \citep{caesar2020nuscenes}} is an AV dataset collected in Boston and Singapore. The autonomous vehicle sensor suite includes 6 12Hz RGB cameras, a 20Hz 32-beam LiDAR, GPS, and an IMU. It also provides rasterized maps featuring drivable areas, stop signs, and pedestrian crossings. 

\textbf{Waymo Open Dataset} is an AV dataset compiled in San Francisco, Phoenix, and Mountain View. The data collection platform incorporates 5 LiDARs, 5 RGB cameras, GPS, and an IMU, with samples collected at 10 Hz. Maps are provided in vectorized form.

\textbf{Data Representation}: We generate OGMs using a ground-segmented LiDAR point cloud. The OGM dimensions are $H \times W = 128 \times 128$ with a \SI[parse-numbers=false]{0.\overline{3}}{\metre} resolution, corresponding to a \SI{42.7}{\metre} $\times$ \SI{42.7}{\metre} grid. We downscale RGB images and maps to the same resolution. During the sequence prediction training phase, we provide 5 past OGMs (\SI{0.5}{\second}) as observations alongside other sensor modalities, and predict for 15 frames (\SI{1.5}{\second}) into the future at \SI{10}{\hertz}. For all open-source datasets, we adhere to the original dataset specification for the train-validation-test split. %We describe the exact LiDAR processing pipeline in \cref{sec:data_details}.

%We show examples of the processed representations in \cref{fig:datasets}.
%While we do not use radar measurements in our current model, they can be readily incorporated into the framework if needed. 

% \input{Figures/datasets}

\subsection{Architecture and Training Details}
\label{sec:training}
\textbf{Architectures}: Our encoder and decoder employ a convolutional network with residual connections, and a latent vector $z\in\R ^{64 \times 4 \times 4}$. The discriminator is multi-scale and multi-patch, following previous work~\citep{isola2017image, wang2018vid2vid, shaham2019singan}. We use a vanilla Transformer implementation provided in PyTorch~\citep{paszke2019pytorch}. 
% During the training process, we utilize a perceptual loss~\citep{zhang2018unreasonable}, KL regularization~\citep{kingma2013auto}, and patch-based adversarial losses~\citep{isola2017image}.
%In \cref{eqn:autoenc}, the KL regularization weight is set to $\beta=4.0$, and the adversarial loss weight to $\gamma=0.01$. 
%We implement the prediction network with a transformer~\citep{vaswani2017attention} with PyTorch~\citep{paszke2019pytorch}.

\textbf{Model Training}: We trained the models using PyTorch Lightning 1.4.2 \citep{falcon2019pytorch, paszke2019pytorch}. We used the AdamW optimizer \citep{loshchilov2017decoupled} with a learning rate of $4 \times 10^{-4}$.
% , $\beta=(0.9, 0.999)$, $\epsilon=10^{-8}$, and $\lambda=0.01$. 
For representation learning, we used four NVIDIA V100 32 GB GPUs. The autoencoders were trained for 80k steps with a batch size of~128. We trained all prediction models on a Nvidia Titan RTX 24 GB GPU with 256 batch size. %We further detail hyperparameters in \cref{sec:implementation_details}.

\subsection{Evaluation}
\label{sec:evaluation}
\textbf{Baselines}: We benchmark against methods commonly used in L-OGM prediction, such as PredNet~\citep{lotter2016deep, itkina2019dynamic} and TAAConvLSTM~\citep{lange2020attention}. Due to their representational similarity to our approach, we also draw comparisons with available SOTA video prediction methods of similar scale like SimVP V2~\citep{tan2022simvp}, PredRNN V2~\citep{wang2022predrnn}, and E3DLSTM~\citep{wang2019eidetic}. We also introduce a baseline that uses the last observed frame as a prediction to assess how effectively the models capture the scene's motion. 
%and Double-Prong ConvLSTM~\citep{toyungyernsub2022dynamics}
% Move to appendix to save space.
%For training, we use open-source implementations provided by OpenSTL~\citep{li2023openstl}. %We provide more details in \cref{sec:baselines_details}.

\textbf{Quantitative Evaluation}: We evaluate all models using the Image Similarity (IS) metric~\citep{birk2006merging} across the \SI{1.5}{\second} and \SI{3.0}{\second} prediction and the accuracy of occupied cells at the end of the \SI{3.0}{\second} rollout. For stochastic predictions, we sample 10 predictions and evaluate the best-performing one. Similar to the mean squared error (MSE) used in trajectory prediction, we are interested in the relative distance errors between observed objects and predicted position of the object but in the discretized space. The IS metric
captures the variations in predicted multi-future agent positions~\citep{lange2020attention}. Commonly used precision-focused metrics, like MSE, penalize a missing moving agent significantly less than a predicted agent's slight positional shift when compared with the ground truth~\citep{lange2020attention}. The IS metric calculates the smallest Manhattan distance between two grid cells with the same thresholded occupancy. Its numerical evaluation is proportional to the discrepancy between predicted and observed positions and is notably high when an agent completely disappears. We also provide an accuracy of occupied cells at the end of the \SI{3.0}{\second} rollout to highlight the critical issue of vanishing moving objects, a common and prohibitive failure for safety-critical applications.

\section{Results}
\label{sec:results}
\subsection{Latent Space Analysis}
\label{sec:latent_analysis}

\input{Figures/reconstruction_images}

The latent space trained in the representation learning stage is essential to facilitate accurate predictions. If an agent in the observed frames is lost during the encoding phase, the prediction network will have difficulties recovering this information, potentially leading to incorrect forecasting. We investigate the reconstruction performance qualitatively and quantitatively. Our framework successfully recovers each sensor modality's observation from their respective low-dimensional representations as visualized in \cref{fig:reconstruction}. We investigate the impact of KL regularization in \cref{tab:kl}. We observe that high KL regularization is needed for the L-OGM's VAE-GAN to generate high quality predictions. 
% We hypothesize that when latent vectors of real OGMs are too scattered around the latent space, any approximation errors will lead to the predicted vectors leaving the latent space representing real occupancy grids and hence lead to the low quality of the predictions.
The KL term induces better regularization and smoothness in the learned latent space, resulting in improved generalization. 
% We hypothesize that any approximation errors during prediction will lead to the predicted latent variables leaving the ground truth data distribution.
We hypothesize that predictions in the latent space will unavoidably incur some inaccuracies, and thus, will not perfectly correspond to the ground truth L-OGM data distribution. This better generalization is important when making predictions in the latent space. We also report the reconstruction performance of camera and map observations. We did not observe significant impact of their regularization on the prediction task.
\vspace{-1.0em}
\begin{table}[h]
    \centering
    \caption{Impact of KL regularization during representation learning on LOPR's performance in terms of reconstruction and a simplified prediction task using the NuScenes dataset. For all metrics, lower is better.}
    \tiny 
    \begin{tabular}{@{}lcccc@{}}
      \toprule
      \textbf{KL} & \textbf{OGM Recon. (IS)} & \textbf{OGM Prediction (IS)} & \textbf{Camera Recon. (MSE)} & \textbf{Map Recon. (MSE)} \\ 
      \midrule
      \num{1e-06} & 3.51 $\pm$ 0.01 & 14.85 $\pm$ 0.72 & \textbf{2.21 $\pm$ 0.04} & \textbf{0.49 $\pm$ 0.04}  \\
      \num{1.0} & \textbf{1.71 $\pm$ 0.01} & 15.90 $\pm$ 0.34 & 2.89 $\pm$ 0.07  & 0.51 $\pm$ 0.06 \\
      \num{4.0} & 1.75 $\pm$ 0.01 & \textbf{13.19 $\pm$ 0.42} & 3.12 $\pm$ 0.06 & 0.53 $\pm$ 0.07 \\
      \bottomrule
    \end{tabular}
  \label{tab:kl}
\vspace{-2.7em} % Add some vertical spacing between the rows
\end{table}

% However, the most largest impact on quality of the predictions has KL regularization of the OGM encoder and decoder pairing on the deterministic prediction task. We observed that some amount of KL regularization is needed to enable high quality predictions in the latent space, as visualized in \cref{fig:kl_intep}. Otherwise, latent vectors of real OGMs are too scattered around the latent space and any prediction errors will lead to the predicted vectors leaving latent vectors representing real occupancy grids.

% We further explore the interpretability of our latent space by interpolating between two OGMs within the same scene sequence. We observe that the interpolated frames have a similar visual appearance to the ground truth OGMs and capture the motion of the environment. In \cref{fig:kl_intep}, we visualize a scene from the test set where the interpolation captured the horizontal motion of an agent.

\subsection{Prediction Task}
\label{sec:pred_task}
\begin{figure}[t]
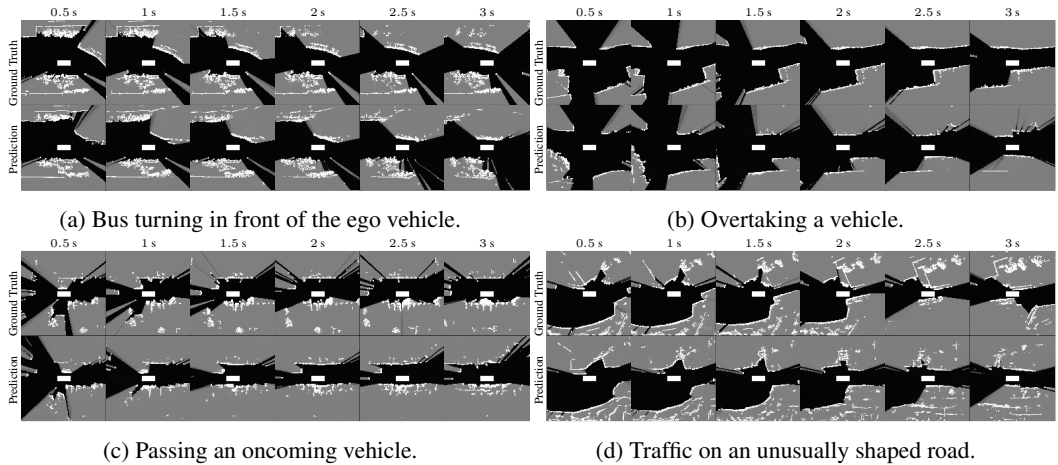

  \centering
  % Top-left figure
  \begin{subfigure}[b]{0.5\textwidth}
    \centering
    \input{Figures/prediction-1_v2.tex}
    \vspace{-1.0em}
    \caption{Bus turning in front of the ego vehicle.}
    \label{fig:top_left}
  \end{subfigure}%
  % Top-right figure
  \begin{subfigure}[b]{0.5\textwidth}
    \centering
    \input{Figures/prediction-2_v2.tex}
     \vspace{-1.0em}
    \caption{Overtaking a vehicle.}
    \label{fig:top_right}
  \end{subfigure}
  
  \vspace{-0.2em} % Add some vertical spacing between the rows
  
  % Bottom-left figure
  \begin{subfigure}[b]{0.5\textwidth}
    \centering
    \input{Figures/prediction-3_v2.tex}
     \vspace{-1.0em}
    \caption{Passing an oncoming vehicle.}
    \label{fig:bottom_left}
  \end{subfigure}%
  % Bottom-right figure
  \begin{subfigure}[b]{0.5\textwidth}
    \centering
    \input{Figures/prediction-4_v2.tex}
     \vspace{-1.0em}
     \caption{Traffic on an unusually shaped road.}
    \label{fig:bottom_right}
  \end{subfigure}
  \vspace{-1.0em} % Add some vertical spacing between the rows
  \caption{\textcolor{black}{Predictions from NuScenes conditioned on L-OGM only. Ego vehicle is moving to the right. Moving objects are realistically propagated in the scene and the details of the static environment are maintained.}}
  %From the top: 1) Oncoming vehicle is passing our ego to the left of it. Our stochastic model is capable of hallucinating a vehicle entering the observable part of the OGM. 2) Ego vehicle is surrounded by traffic. 3) Ego vehicle is passed by a large bus. In all of the scenarios, LOPR correctly propagates the moving objects. 
  \label{fig:prediction}
\vspace{-1.0em} % Add some vertical spacing between the rows
\end{figure}

\begin{figure}[t]
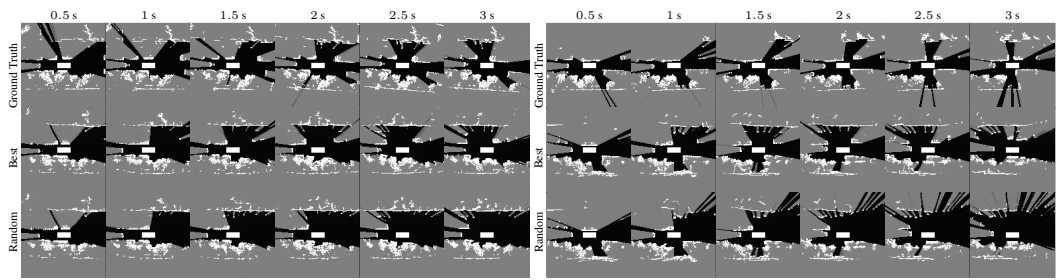

  \centering
  % Top-left figure
  \begin{subfigure}[b]{0.5\textwidth}
    \centering
    \input{Figures/prediction_stochastic-1}
   % caption{Top-left visualization}
    \label{fig:top_left}
  \end{subfigure}%
  % Top-right figure
  \begin{subfigure}[b]{0.5\textwidth}
    \centering
    \input{Figures/prediction_stochastic-2}
    %\caption{Top-right visualization}
    \label{fig:top_right}
  \end{subfigure}
\vspace{-2.0em} % Add some vertical spacing between the rows
\caption{\textcolor{black}{Comparison of best predictions and random predictions from NuScenes conditioned on L-OGM only. Our framework is capable of inferring a previously unobserved oncoming agent entering the L-OGM in both scenes and different variations of static environment.}} %.  Our framework realistically forecasts potential futures of observed agents and reasons over the plausible occluded environments and other agents entering the scenes.
\label{fig:prediction_samples}
\end{figure}

We assess the predictive capabilities of LOPR in \cref{tab:results} and show its predictions in \cref{fig:prediction,fig:prediction_samples}. 
We investigate the influence of stochasticity and the addition of other sensor modalities by evaluating deterministic and stochastic models conditioned only on L-OGMs, and another stochastic model conditioned on all available inputs. 
All LOPR variations notably outperform prior work in L-OGM prediction, as evidenced by the IS and accuracy metrics. 
The improvements become more pronounced with the integration of additional modalities and stochastic modeling.
LOPR is also the only framework that markedly outperforms the simple baseline, which repeats the last observation, demonstrating its ability to capture the scene's dynamics.

In \cref{fig:prediction}, we visualize the predictions rolled out for \SI{3.0}{s} into the future, i.e. beyond the prediction horizon used during training. LOPR produces high-quality, realistic predicted frames, supporting quantitative results.
Dynamic objects are realistically propagated in the scene and the details of the static environment are maintained, unlike prior approaches~\cite{itkina2019dynamic,lange2020attention}. \cref{fig:prediction_samples} presents varied prediction samples from challenging scenarios characterized by partial observability. Each sample captures a realistic plausible future rollout.
Our framework is capable of inferring a previously unobserved agent entering the L-OGM. Our framework is the first, to the best of our knowledge, to successfully generate realistic stochastic OGM predictions conditioned on multi-modal sensor data.

Prior work that uses ConvLSTM-based architectures optimized for the spatiotemporal prediction task in grid cell space~\citep{itkina2019dynamic, schreiber2019long, lange2020attention, toyungyernsub2021double} suffers from lack of sharpness and poor realism in their predictions. For example, occluded cells are often not preceded by occupied cells, making the frames physically incorrect, and forecasted L-OGMs gradually lose important details over the prediction horizon. 
Lack of sharpness eventually leads to the vanishing of moving objects and, hence, missing agents in the predictions~\cite{itkina2019dynamic,lange2020attention} and a low IS metric value. These failures are not present in LOPR's predictions.

% We also demonstrate successful transfer of our LOPR model trained on the NuScenes dataset to our custom robotic dataset in \cref{tab:results,fig:prediction}. The quantitative prediction performance on this dataset is comparable in magnitude to the other datasets, despite the data distribution shift. Thus, LOPR is able to successfully generalize beyond its training data distribution.
%Our generative model learns an expressive latent space for OGM data, enabling the prediction network to make high-quality, realistic predictions. As can be seen from \cref{tab:results} and \cref{sec:latent_analysis}, approximately 30\% of numerical error can be attributed to losses during the encoding stage. The errors occur mainly in the distribution of occupied cells within an object and are, thus, difficult to observe during visual inspection (see \cref{fig:reconstruction}). Despite these losses, the proposed LOPR framework outperforms all OGM prediction baselines in terms of prediction quality (see \cref{tab:results}). 

\begin{table}[h]
    \centering
    \caption{NuScenes and Waymo Open Perception prediction results.}
    \tiny 
    \scalebox{1.2}{
    \begin{tabular}{@{}lcccccccccc@{}}
      \toprule
      \textbf{Model} & \multicolumn{3}{c}{\textbf{NuScenes Dataset}} & \multicolumn{3}{c}{\textbf{Waymo Open Perception Dataset}} \\ 
      & $\textbf{IS}_{5\rightarrow15} (\downarrow)$ & $\textbf{IS}_{5\rightarrow30} (\downarrow)$  & \textbf{Acc.} ($\uparrow$) & $\textbf{IS}_{5\rightarrow15} (\downarrow)$  & $\textbf{IS}_{5\rightarrow30} (\downarrow)$ & \textbf{Acc.} ($\uparrow$ \\
      \midrule
      ED3LSTM~\citep{wang2019eidetic}& 8.25 $\pm$ 0.13 & 18.96$\pm$1.59 & 0.13$\pm$0.01 & 8.97$\pm$0.21 & 41.05$\pm$4.31 & 0.02$\pm$0.01 \\
      TAAConvLSTM~\citep{lange2020attention}& \textbf{6.90$\pm$0.09} & 15.23$\pm$3.94 & 0.14$\pm$0.01 & 6.58$\pm$ 0.42  & 21.96$\pm$2.01 & 0.27$\pm$0.02 \\
      PredRNN V2~\citep{wang2022predrnn}& 13.04$\pm$0.24 & 79.95$\pm$4.64 & 0.03$\pm$0.01 & 11.85$\pm$0.027 & 68.78$\pm$6.06 & 0.11$\pm$0.01 \\
      Sim.VP V2~\citep{tan2022simvp}& 11.86$\pm$0.19 & 54.09$\pm$2.87 & 0.02$\pm$0.01 & 8.77$\pm$0.20 & 48.35$\pm$3.57 & 0.06$\pm$0.01 \\
      PredNet~\citep{lotter2016deep,itkina2019dynamic}& 7.01$\pm$0.14 & 13.89$\pm$0.43 & 0.14$\pm$0.01 & 6.92$\pm$0.61 & 22.26$\pm$1.32 & 0.35$\pm$0.05 \\
      \textcolor{black}{Fixed Last Obs Frame} & \textcolor{black}{11.50$\pm$0.14} & \textcolor{black}{14.41$\pm$0.18} & \textcolor{black}{0.20$\pm$0.01} & \textcolor{black}{10.35$\pm$0.41} & \textcolor{black}{14.74$\pm$0.54} & \textcolor{black}{0.38$\pm$0.01} \\
      \midrule
      Ours \\
      \midrule
      \textcolor{black}{Deterministic (L-OGM)}  & \textcolor{black}{8.09$\pm$0.26} & \textcolor{black}{11.48$\pm$1.38} & \textcolor{black}{0.40$\pm$0.01} & \textcolor{black}{5.92$\pm$0.63} & \textcolor{black}{12.49$\pm$2.05} & \textcolor{black}{0.44$\pm$0.02} & \\
      Stochastic (L-OGM)   & 7.22$\pm$0.10 & 9.78$\pm$0.34 & 0.59$\pm$0.02 & \textbf{5.73$\pm$0.65} & 12.32$\pm$2.34 & \textbf{0.51$\pm$0.07} \\
      \textcolor{black}{Stochastic (All)} & \textcolor{black}{7.01$\pm$0.24} & \textcolor{black}{\textbf{9.16$\pm$0.68}} & \textcolor{black}{\textbf{0.61$\pm$0.01}} & \textcolor{black}{5.85$\pm$0.31} & \textcolor{black}{\textbf{12.18$\pm$2.62}} & \textcolor{black}{\textbf{0.51$\pm$0.01}} \\
      \bottomrule
    \end{tabular}}
  \label{tab:results}
\end{table}

%We believe the impact of sensor modalities will be more siginificant with larger datasets or if pretrained image models, such as X, are applied.
\section{Limitations}
\label{sec:limitations}
Although decoupling the representation learning and prediction stages allows us to achieve high-quality OGM predictions empirically, LOPR relies on a well trained latent space that may not be tuned for prediction. If the latent encoding misses important scene information (such as moving agents), the prediction network will not be able to recover this information and fail to predict the agent. Furthermore, as any learned module in a safety-critical engineering system, like an AV, LOPR would need to be tested rigorously against potential out-of-distribution (OOD) scenarios. LOPR does not currently identify OOD inputs that may cause failures. To enable such capability, we could adopt an approach for epistemic uncertainty estimation as proposed by \citet{itkinacorl2022}.

\section{Conclusion}
\label{sec:conclusion}

In this paper, we proposed an L-OGM prediction framework, LOPR, that decouples representation and prediction learning. LOPR consists of a VAE-GAN-based generative model that learns an expressive low-dimensional latent space of available sensor modalities and a transformer-based stochastic prediction network that operates on the learned latent space. Our experiments show that LOPR achieves SOTA performance, qualitatively and quantitatively outperforming prior L-OGM approaches and video prediction architectures. We also demonstrate that our framework can be easily deployed on any robotic platform equipped with a LiDAR sensor, in contrast to most trajectory prediction approaches, which would require addressing compounding data distribution errors across modules and changes in sensor positioning. We hope that learnings from this framework open new research avenues in the L-OGM prediction field. In future work, we will explore how LOPR can be extended to perform 3D occupancy prediction, be applied to other tasks, such as occlusion inference~\cite{itkinaicra2022} and path planning~\cite{dosovitskiy2017carla}, \textcolor{black}{and serve as a pre-training task for supervised models, such as trajectory prediction.}

\acknowledgments{This project was made possible by funding from the Ford-Stanford Alliance.}

% %===============================================================================

% % no \bibliographystyle is required, since the corl style is automatically used.
%\newpage
\bibliography{main}  % .bib
\newpage
\appendix
\section{Hyperparameteres}
\label{sec:implementation_details}
\subsection{Encoder \& Decoder}
\begin{table}[h]
    \centering
    \caption{Representation Learning Hyperparameters}
    \label{tab:hyperparams_rl}
    \begin{tabular}{@{}lr@{}} %{lX}
        \toprule
        \textbf{Parameter} & \textbf{Value} \\ 
        \midrule
        Architecture & ResNet \\
        Image Size & 1 $\times 128 \times 128$ \\
        Latent Dim L-OGM & $64 \times 4 \times 4$ \\
        Latent Dim Camera & $16 \times 4 \times 4$ \\
        Latent Dim Maps & $16 \times 4 \times 4$ \\
        Channel Multiplier & 2 \\
        Optimizer & Adam(lr=0.0001) \\
        Batch size & 24 per GPU \\
        KL reg $\beta$ & 50.0 \\
        Adv weight $\gamma$ & 1.0 \\
        Path reg & 2.0 \\
        Discriminator R1 reg & 10.0 \\
        GPUS & 4 NVIDIA V100 32 GB \\ 
        \bottomrule
    \end{tabular}
\end{table}
\subsection{Prediction Network}
\Cref{tab:hyperparams_stoch_pred} contains the hyperparameters for stochastic prediction. The prediction network training consists of three stages: deterministic training, low regularization training, and high regularization training. During deterministic training stage, $z_{stoch}$ is sampled from the target distribution. During regularized training, $z_{stoch}$ is sampeld from the inference network.   
\begin{table}[h]
    \centering
    \caption{Stochastic Prediction Hyperparameters}
    \label{tab:hyperparams_stoch_pred}
    \begin{tabular}{@{}lr@{}} %{lX}
        \toprule
        \textbf{Parameter} & \textbf{Value} \\ 
        \midrule
        Architecture & Transformer \\
        $d_{embed}$ & 256 \\
        $N_{enc}$ & 2 \\
        $N_{dec}$ &  1 \\
        $N_{heads}$ & 2 \\
        $d_{feedforward}$ & 128 \\
        Dropout & 0.01 \\ 
        Optimizer & AdamW(lr=0.0004) \\
        Deterministic Epochs & 10 \\
        Low Reg Epochs & 10 \\
        KL reg low $\gamma$ &  0.0001 \\
        KL reg $\gamma$ & 0.001 \\
        GPU & 1 NVIDIA RTX TITAN 24 GB \\
        \bottomrule
    \end{tabular}
\end{table}

\section{Image Similarity Metric}
\label{sec:is}
The Image Similarity metric determines the picture distance function $\psi$ between two matrices $m_1$ and $m_2$ as follows~\cite{birk2006merging}:
\begin{equation}
\begin{split}
    \psi(m_1,m_2) = \sum_{c \in \mathcal{C}} d(m_1,m_2,c) + d(m_2,m_1,c) 
\end{split}
\end{equation}
where
\begin{equation}
\begin{split}
    d(m_1,m_2,c) = \frac{\sum_{m_1[p]=c} \text{min} \{\text{md}(p_1,p_2)|m_2[p_2]=c\}}{\#_c(m_1)}.
\end{split}
\end{equation}

\noindent $\mathcal{C}$ is a set of discretized values assumed by $m_1$ or $m_2$ which are: occupied, occluded, and free. $m_1[p]$ denotes the value $c$ of map $m_1$ at position $p=(x,y)$. $\text{md}(p_1,p_2)=|x_1 - x_2| + |y_1 - y_2|$ is the Manhattan distance between points $p_1$ and $p_2$.  $\#_c(m_1)= \#\{p_1 \mid m_1[p_1]=c\}$ is the number of cells in $m_1$ with value $c$.

\section{Additional Qualitative Results}
\begin{figure}[h]
    \centering
    \centerline{\includegraphics[width=12.0cm]{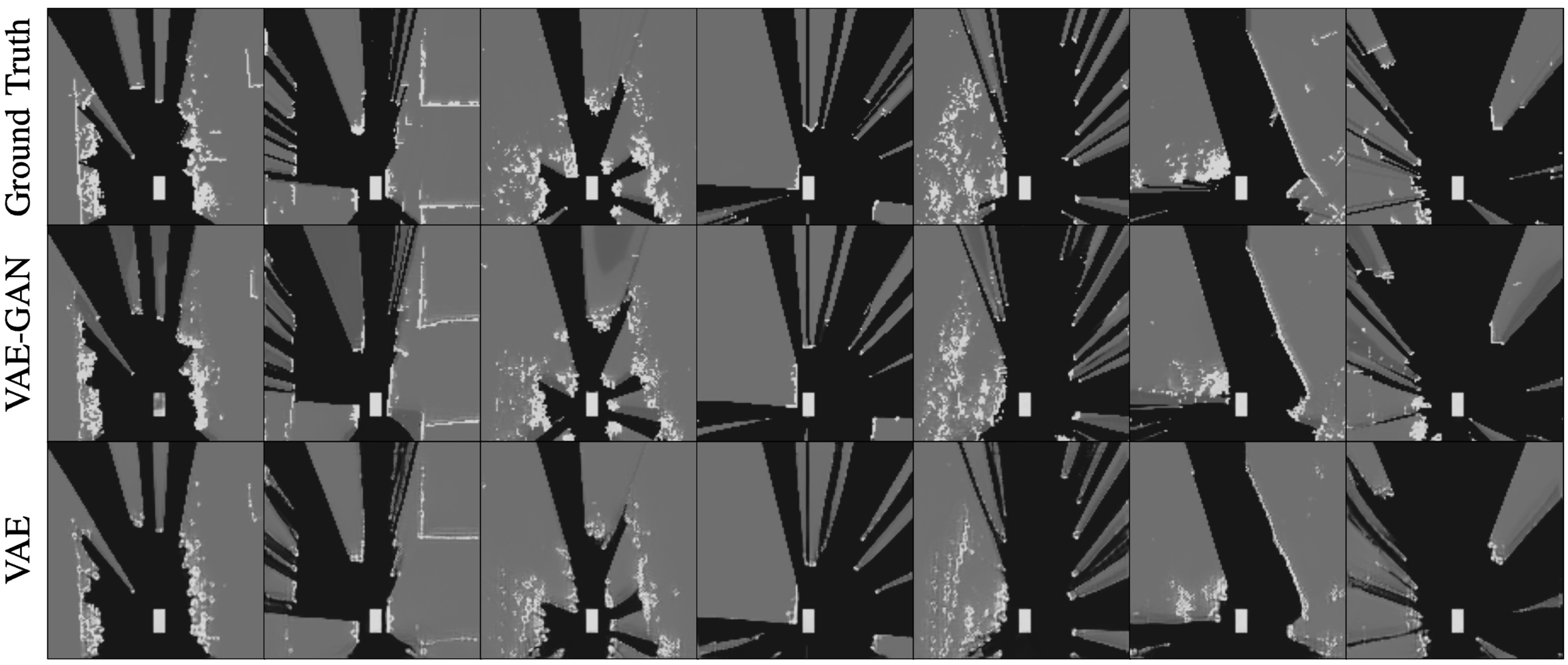}}
    \caption{Comparison between VAE-GAN and VAE on Nuscenes.}
    \label{fig:vae_vaegan_comp}
\end{figure}

\Cref{fig:vae_vaegan_comp} compares VAE-GAN (used in our experiments) and VAE. Samples from VAE-GAN appear more realistic and have sharper occupied cells. Besides the qualitative preferences, the choice of representation learning framework does not translate to significant quantitative differences. 
\end{document}